\documentclass[10pt,journal,compsoc]{IEEEtran}

\ifCLASSOPTIONcompsoc
  \usepackage[nocompress]{cite}
\else
  \usepackage{cite}
\fi

\usepackage{graphicx}
\usepackage{amsmath,amssymb} 
\usepackage[table]{xcolor}
\usepackage{balance}
\usepackage{url} 

\ifCLASSINFOpdf
\else
\fi
\begin{document}
%
\title{Hyperspectral recovery from RGB images using Gaussian Processes}
%
%
%
%

\author{Naveed~Akhtar 
        and~Ajmal~Mian 
\IEEEcompsocitemizethanks{\IEEEcompsocthanksitem N. Akhtar and A. Mian are with the 
Department of Computer Science and Software Engineering (M002), The University of Western Australia, 35 Stirling Highway, CRAWLEY 6009. WA, Australia.\protect\\
E-mail: \{naveed.akhtar, ajmal.mian\}@uwa.edu.au}
\thanks{Manuscript received .......; revised .......}}

%
%

\markboth{Journal of \LaTeX\ Class Files,~Vol.~14, No.~8, August~2015}%
{Shell \MakeLowercase{\textit{et al.}}: Bare Advanced Demo of IEEEtran.cls for IEEE Computer Society Journals}
%



\IEEEtitleabstractindextext{%
\begin{abstract}
We propose to recover spectral details from RGB images of known spectral quantization by modeling natural spectra under Gaussian Processes and combining them with the RGB images. Our technique exploits Process Kernels to model the relative smoothness of reflectance spectra, and encourages non-negativity in the resulting signals for better estimation of the reflectance values.
The Gaussian Processes are inferred in sets using clusters of {spatio-spectrally} correlated hyperspectral training patches. Each set is transformed to match the spectral quantization of the test RGB image. We extract overlapping patches from the RGB image and match them to the  hyperspectral training patches by spectrally transforming the latter. The RGB patches are  encoded over the transformed Gaussian Processes related to those hyperspectral patches and the resulting image is constructed by combining the codes with the original Processes. Our  approach infers the desired Gaussian Processes under a fully Bayesian model inspired by Beta-Bernoulli Process, for which we also present the inference procedure. A thorough evaluation using three hyperspectral datasets demonstrates the effective extraction of spectral details from RGB images by the proposed technique.

\end{abstract}

}

\maketitle

\IEEEdisplaynontitleabstractindextext

%
\IEEEpeerreviewmaketitle

\ifCLASSOPTIONcompsoc
\IEEEraisesectionheading{\section{Introduction}\label{sec:introduction}}
\else
\section{Introduction}
\label{sec:introduction}
\fi

%
%
%
%
\IEEEPARstart{H}{yperspectral} images preserve the fine spectral details  that are lost in RGB images due to the gross quantization of scene radiance.  These details are desirable in numerous imaging applications, 
ranging from Remote Sensing to Medical Imaging and Forensics \cite{TNNLS},  \cite{5762314},  \cite{1410}, \cite{edelman2012hyperspectral}, \cite{Kawakami:2011:HHI:2191740.2191889}, \cite{lu2014medical}, \cite{Zhou_2014_CVPR}.
More recently, spectral characteristics of signals have also been shown to improve performance in multiple Computer Vision tasks, such as  recognition~\cite{Uzair:2013}, \cite{Uzair:2015}, \cite{Zhang:2012:CSP:2071389.2071391};   tracking~\cite{kim2011visual}, \cite{12480}; pedestrian detection~\cite{Hwang_2015_CVPR} and  document analysis~\cite{khan2015automatic}. 
This fact has motivated a wider interest of Computer Vision community in  hyperspectral imaging~\cite{Akhtar:2014a}, \cite{Akhtar:2015}, \cite{Arad2016}, \cite{Dian_2017_CVPR}, \cite{Lanaras:2015}, \cite{OnshotCVPR17}. 

Hyperspectral cameras integrate scene radiance under dozens (or even hundreds) of spectrally well-localized basis functions.
This results in images with a large number of spectral channels, each capturing the scene reflectance in a precise wavelength range.
However, to maintain an acceptable signal-to-noise-ratio of each spectral channel, hyperspectral cameras generally require longer exposure time. 
Moreover, as compared to their color counterparts, present-day hyperspectral cameras severely lack in terms of  spatial resolution~\cite{Akhtar:2015}, \cite{Lanaras:2015}. 
This problem stems directly from the fact that increasing the sensor resolution in hyperspectral cameras  further reduces the photon density at the sensor, which is already limited by the narrow wavelength intervals of the spectral channels.
Owing to the specialized hardware that accounts for these issues, current off-the-shelf hyperspectral cameras are expensive and their use is largely limited to the applications where the temporal and spatial resolution of imaging is not  critical. 

Hyperspectral imaging using computational techniques~\cite{Arad2016}, \cite{goel2015hypercam}, \cite{OhCVPR16}, \cite{Gallini_2017} is an affordable alternate to exploit the power of spectral domain in more common imaging applications, which makes it an interesting research problem. 
Goel et al.~\cite{goel2015hypercam} proposed to estimate up to 17 distinct spectral channels for scenes by augmenting a regular color camera with a time-multiplexed illumination source. 
Oh et al.~\cite{OhCVPR16} used the disparity between the spectral responses of multiple color cameras to compute scene reflectance at the wavelengths other than those of camera spectral channels. 
Similarly, Takatani et al.~\cite{OnshotCVPR17} augmented a conventional camera with a tube of faced reflectors and estimated the scene spectra beyond the sampling wavelengths of camera channels.
Arad and Ben-Shahar~\cite{Arad2016} also demonstrated recovery of higher dimensional spectra from three channel color images. 
They factorized the color image and a set of training hyperspectral images into their corresponding bases and coefficient matrices. Later, the coefficient matrix of the color image is used with the hyperspectral basis to construct a hyperspectral image. 

The technique proposed by Arad and Ben-Shahar\cite{Arad2016} is unique in the sense that it does not require any external hardware/illumination source besides a single conventional camera. Inspired by the paramagnetic nature of this paradigm, we also propose to recover spectral details from a single RGB camera. However, in contrast to \cite{Arad2016}, we mainly focus on the physical attributes of reflectance spectra for their accurate recovery.  It is a common knowledge that naturally occurring reflectance spectra are relatively smooth functions of wavelength~\cite{6555921}. Gaussian Processes~\cite{rasmussen2006gaussian} can explicitly model such smoothness using Kernels, which makes them an ideal tool for spectral recovery. Similarly, self-similarity of spatial patterns and spectral sources within a scene~\cite{Dian_2017_CVPR}, and non-negativity of  reflectance are other physical attributes that we exploit to construct hyperspectral signatures  from RGB images.

\begin{figure*}[t] 
   \centering
   \includegraphics[width=5.5in, height =2.8in]{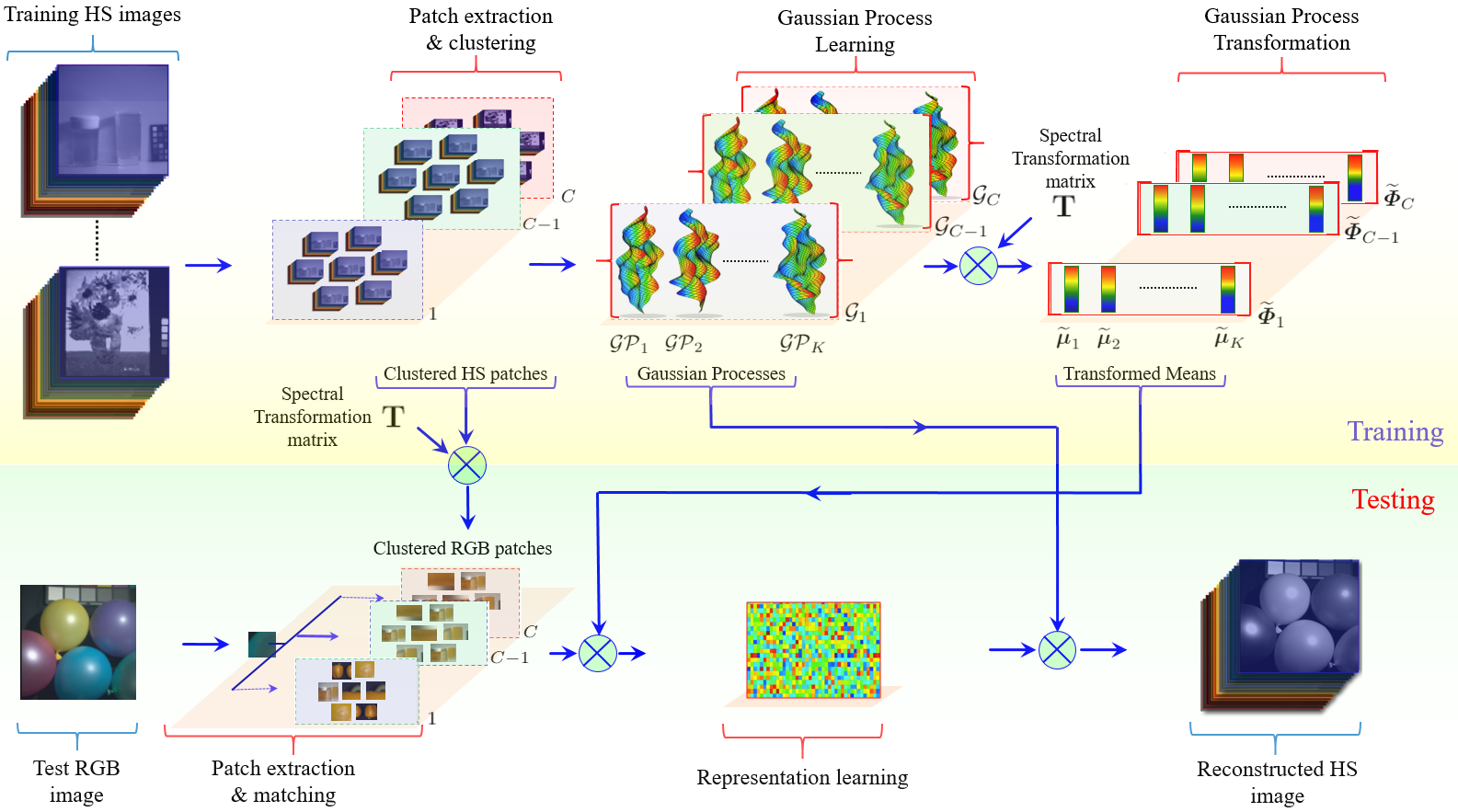} 
   \caption{Schematics of the proposed approach: \emph{Training} - {Spatio-spectral} patches from  training Hyperspectral (HS) images are extracted and clustered. A set of Gaussian Processes is inferred for each cluster under the proposed representation model. These Processes are transformed to match the spectral quantization of the RGB image. \emph{Testing} - RGB image patch is extracted and  matched with the RGB transformations of the HS clusters. The transformed Gaussian Process Means for the matched cluster are used to represent the patch. The representation codes are combined with the original Gaussian Processes to construct the desired HS image. } 
   \label{fig:Schema}
   \vspace{-3mm}
\end{figure*}


Concretely, we propose to learn multiple sets of Gaussian Processes from training hyperspectral image patches.
For the overall approach, an element (i.e.~a Gaussian Process) in a given set signifies a distribution over a spectral source in the training data.
{We can expect hyperspectral image regions containing similar materials/spectral-sources to be spatio-spectrally similar.}
Therefore, we use clusters of {spatio-spectrally}  correlated patches to learn their corresponding sets (see Fig.~\ref{fig:Schema}).
We transform the Mean parameters of the inferred Gaussian Processes to match the spectral quantization of the RGB image.
The transformed Means are used to encode overlapping patches of the RGB image. To ensure effective coding, we first match an RGB patch to the RGB transformations of the hyperspectral training patches, and use the Means corresponding to the right cluster. The codes of RGB patches are combined with the original Gaussian Processes to estimate the detailed spectra of the scene.

We propose to learn the desired Gaussian Processes under a representation model inspired by  Beta-Bernoulli Process~\cite{paisley2009nonparametric}. However, instead of using simple multi-variate Gaussian distributions in the model~\cite{paisley2009nonparametric}, we employ  Gaussian Processes as the base measure with appropriate Kernels for spectral modeling. We additionally place Gamma hyper-prior over the Kernel parameter for its automatic adjustment to the training data. Moreover, we also encourage non-negativity in the inferred signals for modeling the reflectance values.
Our Bayesian representation model is fully conjugate, which allows us to  analytically derive the posterior distribution expressions for inferring the desired Gaussian Processes. We use Gibbs sampling as the inference technique and provide its sampling equations. The approach is thoroughly evaluated by recovering the hyperspectral images from RGB images of more than 35 Indoor, Outdoor and Object scenes from three different hyperspectral databases - CAVE database~\cite{CAVE_0293}, iCVL database~\cite{Arad2016} and the Harvard dataset~\cite{chakrabarti2011statistics}. 
Experimental results demonstrate the effectiveness of the proposed approach, achieving up to $30.75\%$ reduction in the RMSE value for the spectral recovery as compared to~\cite{Arad2016}.

The article is organized as follows. In Section~\ref{sec:RW}, we review the literature related to the proposed approach. The underlying problem is formalized in Section~\ref{sec:PF} and the possibility of recovering spectral details from RGB signals is analyzed in the light of the existing literature in Section~\ref{sec:Extract}. In Section~\ref{sec:PA}, we present the proposed approach, whereas its  implementation details  are discussed in Section~\ref{sec:ImpD}. Experimental  evaluation is presented in Section~\ref{sec:Exp}, and the article concludes in Section~\ref{sec:Conc}.  We also provide the details on the joint probability distribution of the proposed model and the derivations of Gibbs sampling equations for our approach in Appendices (supplementary material). 

\vspace{-3mm}
\section{Related Work}
\label{sec:RW}
Hyperspectral imaging systems have been in use in Remote Sensing platforms for nearly three decades~\cite{solomon1985imaging}.
Early hyperspectral systems such as NASA's AVIRIS~\cite{Green1998227} acquired images using the `whisk broom' scanning method that uses a set of mirrors and fiber optics to re-direct the incoming light to a bank of spectrometers. 
More recent systems use the `push broom' scanning strategy~\cite{james2007spectrograph} that employs dispersive optical elements and light sensitive sensors to obtain a hyperspectral image in a line-by-line manner. 
For the on-ground applications, hyperspectral images are commonly acquired with the help of interchangeable or tunable  narrow band filters~\cite{miller1995multispectral}, \cite{tominaga1996multichannel}.
{However, with the exception of a few, e.g.~coded aperture snapshot spectral imager~\cite{kittle2010}; most of the specialized hardware based strategies fall short of performing}  hyperspectral imaging  with the spatial and temporal resolution  that is generally required in the common Computer Vision tasks.
This fact has led to a considerable recent interest of researchers in the  computational methods for hyperspectral imaging. 

To construct the 3D hyperspectral data cube, Computed Tomography Imaging Spectrometers project the spatio-spectral signals on to 2D sensors using diffraction grating, and computationally estimate  hyperspectral images from these projections~\cite{a},  \cite{c}, \cite{b}. 
Whereas effective hyperspectral cubes can be constructed with this technique, the methods require both specialized hardware as well as considerable post-processing to achieve these results. 
Furthermore, the spatial resolution of the resulting cube is again limited by the used sensor. {Brady} and Gehm~\cite{d}, and later Gehm et al.~\cite{e} used the techniques from compressive sensing to improve on the sensor limitations.
Nevertheless, the specialized image acquisition setup and the significant post-processing remains the bottleneck in the widespread use of the Computed Tomography Imaging Spectrometers for hyperspectral imaging.

Hyperspectral fovea systems~\cite{f}, \cite{g} have also been proposed with an aim to acquire real-time hyperspectral images with less complex post-processing.
These systems capture a high resolution color image along the spectral information of the central region in the image.
Although useful, these systems are more suitable for the applications where occasional hyperspectral  sampling of specific regions is required instead of obtaining the spectral details of the whole  area~\cite{Arad2016}. 
Another direction of computational techniques related to the high resolution hyperspectral imaging~\cite{Akhtar:2014a}, \cite{akhtar2016hierarchical}, \cite{Kawakami:2011:HHI:2191740.2191889},  \cite{Lanaras:2015} uses matrix factorization to compute a spectral basis from a low resolution hyperspectral image and uses it to augment the spectral dimension of the color image of the same scene.
Whereas these methods have to shown impressive reconstruction of hyperspectral cubes, the assumption of availability of the low resolution hyperspectral image can only be satisfied by the specialized hyperspectral hardware. 

A number of computational methods for extracting the spectral information of the scene using color imaging also rely on active scene illumination.
For instance, D'Zmura~\cite{h} estimated the reflectance of the scene with the help of illumination patterns  with independent spectral characteristics. 
Their approach recovers the scene reflectance by finally estimating the coefficients of a linear representation model.
In a similar approach, Park et al.~\cite{i} reconstructed hyperspectral frames of a video using a multiplexed illumination setup that combines different color LEDs to illuminate the scene. 
Chi et al~\cite{j} estimated the spectral reflectance of the scene by placing a set of wide band filters in front of the light source. 
A DLP projector along a high-speed camera was used by Han et al.~\cite{k} to recover the scene reflectance. 
Moreover, a time multiplexed illumination was also employed by Goel et al.~\cite{goel2015hypercam} to estimate the scene reflectance across 17 spectral channels at 9 frames per second. 
Parmer et al.~\cite{l} showed successful estimation of 31 spectral channels with the help of five time-multiplexed LED sources. 
The method images the same scene multiple times to compute the target 31 spectral channels.

Methods for scene reflectance estimation inspired by kaleidoscope geometry also appear in the literature.
For instance, Han and Perlin~\cite{m}  used a tapered kaleidoscope to estimate the bi-directional texture reflectance of a surface.
Manakov et al.~\cite{n} proposed a reconfigurable camera ad-on based on the principle of kaleidoscope for a high dynamic range multi-spectral and light field imaging.
A similar approach has been adopted by Takatani et al.~\cite{OnshotCVPR17} who used faced color filters in a kaleidoscopic configuration for multiple reflections of the incoming light to the sensor.
The different spectra resulting from different reflections are exploited in their approach to computationally estimate the scene reflectance at wavelengths other than the central wavelengths of the spectral sensitivity of the camera.

{Along the lines of the technique proposed in this work, there have  been previous attempts to directly recover spectral details of scenes using RGB cameras. For instance, Nguyen et al.~\cite{nguyen2014training} leveraged RGB white-balancing for scene illumination normalization, and recovered the scene reflectance  using color images and a radial basis function network trained on hyperspectral data. Robles-Kelly~\cite{Antonio2015} extracted spectral and convolutional features from training hyperspectral data by employing sparse coding and used those features to construct hyperspectral images from RGB images. Whereas material properties are considered in \cite{Antonio2015}, the method does not take any advantage of the intrinsic properties of natural spectra. 
A closely related method to our approach is proposed by Arad and Ben-Shahar~\cite{Arad2016}, that learns a hyperspectral basis from training data and transforms it to sparsely encode RGB images. The RGB sparse codes are then combined with the original basis for the spectral recovery of scene reflectance.}

{The technique of Arad and Ben-Shahar closely relates to our previous work~\cite{Akhtar:2014a}, except that it uses a corpus of separately acquired training data instead of a low resolution hyperspectral image of the scene to learn the hyperspectral basis. This eliminates the need of hyperspectral hardware during the testing phase.  
Galliani et al.~\cite{Gallini_2017} exploited hyperspectral training data in the context of deep neural networks~\cite{Deep} to learn a mapping from three to multiple spectral channels. Later, Wu et al.~\cite{Aeschbacher} argued in favor of hyperspectral reconstruction using shallow methods, e.g.~\cite{Arad2016} as opposed to the deep networks  employed by Galliani et al.~\cite{Gallini_2017}. Based on a method exploiting adjusted anchored neighborhood regression for super-resolution~\cite{A+}, Wu et al.~\cite{Aeschbacher} also proposed an improvement over~\cite{Arad2016}. 
Another approach that exploits neural networks for spectral recovery is proposed by Alvarez-Gila et al.~\cite{Gila2017ICCVW} that uses Generative Adversarial Networks~\cite{GANs} to learn a mapping from three to multiple spectral channels. }

{Whereas the above methods relate to the proposed approach, our technique  particularly emphasizes on exploiting the intrinsic attributes of spectral signatures for hyperspectral recovery in a principled manner - in which the existing approaches fall considerably short. It is noteworthy that currently there is a growing interest in the research community for the problem of hyperspectral recovery from RGB images. Very recently, the first on-line challenge on spectral reconstruction from RGB images  has also been introduced at  the NTIRE workshop (\url{http://www.vision.ee.ethz.ch/ntire18/}) conducted in conjunction with CVPR 2018. The challenge required spectral recovery from RGB images constructed by applying known and unknown spectral response functions to the ground truth hyperspectral images. Considering the interest of research community, a principled approach for spectral recovery from RGB images is expected to appropriately guide this emerging direction.  }

\section{Problem formulation}
\label{sec:PF}
We denote an RGB image by ${\bf Y} \in \mathbb R^{M \times N \times l}$, where $ l = 3$ spectral channels. From ${\bf Y} $, we aim to construct a hyperspectral image ${\bf Y}^h \in \mathbb R^{M\times N \times L}$ of the same scene, where $L> l$. From one perspective, our objective is to computationally improve the \emph{spectral} resolution of an RGB image. In this work, we limit this improvement to $10\times$ by letting  $L \leq 10 \times l$. For all practical purposes, we can assume a mapping function $\mathcal M (.): \mathbb R^{M \times N \times L} \to \mathbb R^{M \times N \times l}$ such that ${\bf Y} = \mathcal M({\bf Y}^h)$. This function integrates the spectral channels of the desired hyperspectral image to form the RGB image.  

Let ${\bf y}^h \in \mathbb R^L$ denote a pixel of the hyperspectral image. Due to multiple scatterings of  light and presence of intimate material mixtures in a  scene, this pixel is often formed as a combination of  the spectral signatures of various materials in the scene. 
Considering all the distinct spectral signatures of the scene materials to be present in a matrix  $\boldsymbol{\Phi} \in \mathbb R^{L \times K}$, we may represent a hyperspectral image pixel as  ${\bf y}^h = \boldsymbol{\Phi} \boldsymbol\alpha$, where $\boldsymbol \alpha$ is the coefficient vector in $\mathbb R^K$.
This representation model is commonly employed in hyperspectral unmixing~\cite{myTGRS1}, \cite{6555921}. 
Due to the existence of the mapping function $\mathcal M(.)$, we can further represent the corresponding pixel of the RGB image as ${\bf y} = {\bf T}~\boldsymbol{\Phi} \boldsymbol\alpha$, where ${\bf T} \in \mathbb R^{l \times L}$ is the spectral transformation between the target hyperspectral image and the acquired RGB image. In this work, we consider this  transformation matrix to be known \emph{a priori}.
We note that prior knowledge of ${\bf T}$ is a common assumption in the image fusion based hyperspectral super-resolution literature~\cite{Akhtar:2014a}, \cite{Kawakami:2011:HHI:2191740.2191889}, \cite{Lanaras:2015} and it can be estimated using the spectral response of the cameras.

From the above, we can see that with the help of correctly computed signatures in $\boldsymbol\Phi$ one can hope to recover a much more detailed spectral response of the scene at image pixel level. Assuming a hyperspectral training prior, we aim to estimate the desired signatures using Gaussian Processes that account for the physical attributes of natural spectral, such as their relative smoothness and non-negativity of reflectance values.

\section{Spectral recovery from RGB signals}
\label{sec:Extract}
On the surface, the attempt to recover $\sim\!\!30$-dimensional signatures from three dimensional RGB signals may appear over ambitious.
Arguably, the problem is severely under-constrained because the spectral quantization process of the RGB cameras suffers from the phenomenon of metamerism~\cite{Palmer99a} - perceived/sensed matching of the colors that do not actually match based on the  differences in their spectral power distribution. 
This means, multiple spectral signatures (metamers) can exist for a given RGB pixel. Thus, it is interesting to ask that to what extent hyperspectral signals can really be recovered from the RGB signals?   

Arad and Ben-Shahar~\cite{Arad2016} strongly argued that under certain conditions the detailed spectral information from  RGB signals is indeed recoverable.
The conditions they stipulate for that purpose are as follows. (1)~It is required that the set of spectral signals encountered by the sensor is actually confined to a relatively lower dimensional manifold within the high dimensional space of the possible hyperspectral signals, and (2) the frequency of the metamers in that lower  dimensional manifold is low. If these conditions hold, the RGB signals may  be able to reveal much finer spectral details of the scene than what has been recorded by the sensor.  

Existing studies~\cite{cohen1964dependency}, \cite{maloney1986evaluation}, \cite{parkkinen1989characteristic} have consistently shown that the reflectance spectra of materials are generally well represented by a small number of principal components. 
{Parkinnen et al.~\cite{parkkinen1989characteristic} demonstrated that only eight characteristic spectra are needed to achieve a good representation of 1,257 reflectance spectra of the chips in the Munsell Book of Color - Matte Finish Collection~\cite{cohen1964dependency}.}
Haderberg~\cite{hardeberg2002spectral} computed the effective dimension of the reflectance signatures of a large variety of color pigments to be  between 13 to 23. 
More recently, Chakrabarti and Zickler~\cite{chakrabarti2011statistics} also showed that the first 20 principle components  of their sampled hyperspectral image patches of more than seventy Indoor and Outdoor scenes were able to account for $99\%$ sample variance. Thus, a strong evidence already exists in favor of the above mentioned condition~(1).
In regards to condition~(2), computing the relative frequency of the metameric pairs, Foster et al.~\cite{foster2006frequency} found that this value is usually as low as $10^{-4}$ to $10^{-6}$ for the natural scenes. 
From the perspective of recovering spectral details from  RGB images, it indicates a high probability of recovering the correct metamer from the sensed signal.
Hence, correct recovery of a hyperspectral image from its RGB counterpart indeed seems possible.
Nevertheless, availability of a useful prior over the hyperspectral image is imperative.

In this work, we make the practical assumption that this prior is available in the form of hyperspectral image training data.
Although hyperspectral cameras are currently not as popular as their RGB counterparts, public hyperspectral image databases have started emerging in the recent years.
For instance, Arad and Ben-Shahar~\cite{Arad2016} collected one hundred hyperspectral images of urban, rular, indoor and plant-life scenes with a Specim PS Kappa DX4 camera. Chakrabarti and Zickler~\cite{chakrabarti2011statistics} made public 50 Outdoor and 27 Indoor scene hyperspectral images. The hyperspectral imaging laboratory at the University of Western Australia has also made  hyperspectral image databases for Human Faces~\cite{Uzair:2015} and Documents~\cite{khan2015automatic} publicly available.
Another example of a well-known public database of hyperspectral images of daily life  objects is the CAVE database~\cite{CAVE_0293}. 

\section{Proposed approach}
\label{sec:PA}
Our approach recovers the spectral details from RGB images by inferring multiple sets of Gaussian Processes using relevant hyperspectral training data and meticulously combining them with the RGB images.
We exploit the Kernels of Gaussian Processes to model the relative smoothness~\cite{6555921}, \cite{xing2012dictionary} of naturally occurring spectra. Moreover, the approach also accounts for the non-negativity of reflectance values and the self-similarity of hyperspectral image patches~\cite{Dian_2017_CVPR} for accurate spectral recovery.  
The schematics of the proposed approach is illustrated in Fig.~\ref{fig:Schema}. The overall approach can be divided into training and testing phases. We provide details on each of these phases below.

\subsection{Training}
At the heart of the training phase is learning of Gaussian Processes ($\mathcal {GP}_i \subseteq \mathbb R^L, \forall i $) that represent hyperspectral signatures under a Bayesian model. 
For clarity, we defer the details on the proposed model and its inference scheme to the Sections to follow, and first summarize the remainder of the training phase. 
To learn the Gaussian Processes, the approach pre-processes the training hyperspectral data  by  extracting $p \times p \times L$-dimensional patches from the  images and arranging them in `$C$' clusters using K-Means algorithm.
Processing hyperspectral images in terms of clustered patches is advantageous for our approach for two main reasons.
Firstly, due to their similarity, the signals in each cluster are likely confined to an even  lower dimensional manifold within the signal space.  
Secondly, within each cluster the frequency of the metamers~\cite{Palmer99a} is naturally reduced because the clusters are generally composed of the same/similar spectral sources (e.g. patches corresponding to `grass'  cluster well together).
From Section~\ref{sec:Extract}, we know that these properties of spectral signatures  are conducive for the accurate recovery of the spectral details from RGB images, which benefits the overall approach.

We infer a set $\mathcal G_c = \{ \mathcal{GP}_1, \mathcal{GP}_2,...,\mathcal{GP}_K\} \subseteq \mathbb R^L$ of `$K$'  Gaussian Processes to represent the $c^{\text {th}}$ cluster of the training data.
According to our model, the Gaussian Processes signify the spectral signal sources in the training data.
Hence, we embed the smoothness of spectral signatures into these Processes using their Kernels.
Moreover, we also use positive Mean priors in our model to incorporate non-negativity of reflectance values and the proportions in which the spectral signals generally mix to form image pixels~\cite{1410}. 
With the help of the Bayesian inference (see Section~\ref{sec:Infer}) over the proposed model we learn posterior distributions for the Gaussian Processes.
The Mean parameters of the learned distributions are then transformed to match the spectral quantization of the RGB camera\footnote{The Kernels are not transformed because they are not required in the subsequent processing of the data.}.
Fig.~\ref{fig:Schema} shows the transformed parameters for the $c^{\text {th}}$ cluster as a matrix $\widetilde{\boldsymbol\Phi}_c = [\widetilde{\boldsymbol\mu}_1, \widetilde{\boldsymbol\mu}_2,..., \widetilde{\boldsymbol\mu}_K] \in \mathbb R^{l \times K}$, where $\widetilde{\boldsymbol\mu}_k \in \mathbb R^l$ is the transformed Mean of the $k^{\text{th}}$ Gaussian Process for the cluster.
This matrix is later used in the testing phase of the approach. 


\subsubsection{Representation Model}
\label{sec:RepMod}
To infer the Gaussian Processes we propose to represent the $i^{\text {th}}$ hyperspectral pixel ${\bf y}^h_i \in \mathbb R^L$ as follows:
\begin{align}
\label{eq:mod}
{\bf y}_i^h &= \boldsymbol\Phi \boldsymbol\alpha_i + \boldsymbol\epsilon,\\
\nonumber \boldsymbol\varphi_k  &\sim \mathcal {GP}_k(\boldsymbol\varphi_k | \boldsymbol\mu_{k_o}, \boldsymbol\Sigma_{k_o}),\\
\nonumber \boldsymbol\Sigma_k(\varsigma_a, \varsigma_b) &= \frac{1}{\eta_k} \exp \left(\frac{-|\varsigma_b - \varsigma_a|}{2\ell^2}\right),\\
\nonumber \eta_k&\sim \text{Gam}(\eta_k | a_o, b_o), \\
\nonumber {\boldsymbol\alpha_i} &= {\bf z}_i \odot {\bf s}_i,\\
\nonumber {z}_{ik} &\sim \text{Bern} ({z}_{ik} | \pi_{k_o}),\\
\nonumber \pi_{k} &\sim \text{Beta}(\pi_{k}|{c_o}/{Q}, {d_o(Q-1)}/{Q}),\\
\nonumber s_{ik} &\sim \mathcal N( s_{ik} | \mu_{s_{iko}}, \lambda_{s_o}^{-1}),\\
\nonumber \lambda_{s} &\sim  \text{Gam}(\lambda_{s} | e_o, f_o), \\
\nonumber \boldsymbol\epsilon &\sim \mathcal N(\boldsymbol\epsilon_i  | {\bf 0}, \lambda^{-1}_{\epsilon_o} {\bf I}_L )\\
\nonumber \lambda_{\epsilon} &\sim \text{Gam}(\lambda_{\epsilon} | g_o,h_o).
\end{align}

In the above model, $\boldsymbol\Phi \in \mathbb{R}^{L \times K}$ denotes the matrix composed of a  set of (possibly over-complete) basis vectors $\boldsymbol\varphi_{k \in \{1,2,...,K\}} \in \mathbb R^L$ to represent the  hyperspectral signatures. The vector $\boldsymbol\epsilon \in \mathbb R^L$ models  the Gaussian noise and $\boldsymbol\alpha_i \in \mathbb R^K$ is the coefficient vector for the representation.
We place a separate Gaussian Process prior $\mathcal{GP}_k (.)$ over each of the basis vectors, with the Mean $\boldsymbol\mu_{k_o} \in \mathbb R^L$ and the Kernel $\boldsymbol\Sigma_{k_o} \in \mathbb R^{L\times L}$. In Eq.~(\ref{eq:mod}) and the text to follow, we use the subscript `$o$' to distinguish the parameters of the \emph{prior} probability distributions.
Based on the analogy between the Beta-Bernoulli Process~\cite{paisley2009nonparametric} underlying our model and the Dictionary Learning paradigm~\cite{5714407}, we compute $\boldsymbol\mu_{k_o}$ using the Online Dictionary Learning method proposed by Mairal et al.~\cite{mairal2009online}.
We provide details on this computation in Section~\ref{sec:prior}. 
To promote the spectral correlation between the two consecutive channels of the hyperspectral signals at wavelengths $\varsigma_a$ and $\varsigma_b$ (in nano-meters), we used an exponential Kernel $\boldsymbol\Sigma_k(\varsigma_a, \varsigma_b)$ for the Gaussian Processes.
For the Kernel, the parameter `$\ell$' loosely corresponds to  the number of the consecutive spectral channels of ${\bf y}^h_i$ that the Kernel is able to influence through extrapolation. 
The value $\eta_k^{-1}$ determines the average distance of the underlying function from the Mean, which makes $\eta_k$ the `precision' of the Kernel. 
Therefore, we let the training data itself  determine the value of $\eta_k$ for the proposed model by placing a non-informative Gamma hyper-prior over this parameter. In Eq.~(\ref{eq:mod}), this is denoted by the Gamma distribution Gam(.) with parameters $a_o$ and $b_o$. The parameters in the argument of the exponential function mainly depend on the hardware specifications of the hyperspectral sensor. Therefore, no further hyper-priors are placed  over these parameters. 

Following the formulation of the weighted Beta-Bernoulli Process~\cite{paisley2009nonparametric}, we model the coefficient vector $\boldsymbol\alpha_i$ as the Hadamard product of a binary vector ${\bf z}_i \in \mathbb R^K$ and a Gaussian vector ${\bf s}_i \in \mathbb R^K$. 
The $i^{\text{th}}$ coefficient $z_{ik}$ of  ${\bf z}_i$ is drawn from a Bernoulli distribution Bern(.) with the parameter $\pi_{k_o}$.
Moreover, a conjugate Beta prior, i.e. Beta(.) is placed over $\pi_k$ with the hyper-parameters $c_o, d_o$ and a scaling factor $Q$.
The coefficients $s_{ik}$ of ${\bf s}_k$ are sampled from a Gaussian distribution with the mean $\mu_{s_{iko}}$ and the precision $\lambda_{s_o}$.
Notice that, in this formulation the basis vectors $\boldsymbol\varphi_{k}, \forall k$ are associated with the signals ${\bf y}_{i}^h, \forall i$ under $K$ Bernoulli distributions that govern the support (i.e. ${\bf z}_{i}, \forall i$) of the coefficient vectors. Zhou et al.~\cite{zhou2009non} noted that in Beta-Bernoulli formulation, the Bernoulli parameters automatically adjust to account for the intrinsic sparsity of the signals during the Bayesian inference, thereby generally rendering the coefficient vectors  $\boldsymbol\alpha_i$, $\forall i$ to become sparse.    
Taking advantage of this observation, we estimate the Mean parameters of the prior distributions over ${\bf s}_i, \forall i$ by computing  sparse representations~\cite{tibshirani1996regression} of the training signals over the Mean parameters of the Gaussian Process priors.
The details on this computation are also provided in Section~\ref{sec:prior}.
We further place a non-informative Gamma hyper-priors over the precision $\lambda_s$ of $s_{ik}$ to automatically tune the value of this parameter to the training data.

The model assumes white noise with precision $\lambda_{\epsilon}$ that is sampled from a Gamma distribution with the hyper-parameters $g_o$ and $h_o$. 
We note that the proposed model significantly differs from the original proposal of the Beta-Bernoulli Process~\cite{paisley2009nonparametric} in that (1) it uses Gaussian Processes with explicit Kernels instead of using the standard multivariate Gaussian distributions and (2) it uses non-zero Mean priors over the basis vectors as well as the coefficient vectors to promote non-negativity in the representation. We provide the analytical expression for the joint probability distribution over the training data defined by the proposed model in the Appendix~A (supplementary material).




\subsubsection{Bayesian Inference}
\label{sec:Infer}
Our approach uses model (\ref{eq:mod}) to learn posterior distributions over its parameters for hyperspectral data representation. We use Gibbs sampling technique~\cite{Bishop:2006:PRM:1162264} to infer those distributions. Being fully conjugate, the model allows us to derive analytical expressions for the posterior distributions. 
The inference process uses those expressions to iteratively draw samples from the  respective probability distributions, and converge to the final posterior distributions. 
Below, we provide the analytical expressions used by our Gibbs sampler. 
The sampler sequentially draws samples from each of these distributions and keeps iterating until convergence. 
To avoid cluttering, we only provide the final expressions of the distributions below. The details of arriving at these expressions are separately given in Appendix~B.

In the equations below, $||.||_2$ denotes the $\ell_2$-norm of a vector and $ {\bf y}^h_{i_{\varphi_k}}$ is the contribution of the $k^{\text{th}}$ basis vector in representing ${\bf y}_i^h$. That is, 
$
{\bf y}^h_{i_{\varphi_k}} = {\bf y}_i^h - \boldsymbol\Phi({\bf z}_i \odot {\bf s}_i) + \boldsymbol\varphi_k  (z_{ik} \odot  s_{ik})
$. 
For brevity, we also let  $\widetilde{\boldsymbol\Sigma}_k = \exp\left(\frac{-|\varsigma_b - \varsigma_a |}{2\ell^2} \right)$.\\

\noindent{\bf Sample $\boldsymbol\varphi_k$:} from  $\mathcal {GP} (\boldsymbol\varphi_k | \boldsymbol\mu_k, \widehat{\boldsymbol\Sigma}_k)$, where
\begin{align*}
\widehat{\boldsymbol\Sigma}_k &= \left(\boldsymbol\Sigma_{k_o}^{-1} + \lambda_{\epsilon} \sum\limits_{i=1}^{N} (s_{ik}z_{ik})^2\right)^{-1}, \\
\boldsymbol\mu_k &=  \widehat{\boldsymbol\Sigma}_k \left( \lambda_{\epsilon}  \sum\limits_{i = 1}^{N}(s_{ik}z_{ik}) {\bf y}_{i_{\varphi_k}}^h +  \boldsymbol\Sigma^{-1}_{k_o} \boldsymbol\mu_o  \right).
\end{align*}

\noindent{\bf Sample $\eta_k$:} from Gam$(\eta_k| a, b)$, where
\begin{align*}
a = a_o + \frac{L}{2} , \hspace{2mm} b = b_o + \frac{1}{2} (\boldsymbol\varphi_k - \boldsymbol\mu_k)^{\intercal} \widetilde{\boldsymbol\Sigma}_k (\boldsymbol\varphi_k - \boldsymbol\mu_k).
\end{align*}

\noindent{\bf Sample $z_{ik}$:} from $\text{Bern}\Big( z_{ik}| \frac{\pi_{k_o}\xi}{1-\pi_{k_o} + \xi \pi_{k_o} }\Big)$, where
\begin{align*}
\xi = \exp\Big( - \frac{\lambda_{\epsilon}}{2}(\boldsymbol\varphi_k^{\intercal} \boldsymbol\varphi_k s_{ik}^2  - 2s_{ik}{\bf y}_{i_{\varphi_k}}^{h \intercal}\boldsymbol\varphi_k) \Big).
\end{align*}

\noindent{\bf Sample $s_{ik}$:} from $ \mathcal N (s_{ik} | \mu_{s_{ik}}, \lambda_s^{-1})$, where
\begin{align*}
\lambda_s &=  \lambda_{s_o} +  \lambda_{\epsilon} z_{ik}^{2} \boldsymbol\varphi_k^{\intercal} \boldsymbol\varphi_k,\\
\mu_{s_{ik}}& = \lambda_s^{-1} \Big( \lambda_{\epsilon} z_{ik} \boldsymbol\varphi_k^{\intercal} {\bf y}_{i_{\varphi_k}}^h +\lambda_{s_o}\mu_{s_{iko}}\Big).
\end{align*}   

\noindent{\bf Sample $\pi_k$:} from $\text{Beta}\left( \pi_k | c, d \right)$, where
\begin{align*}
c = \frac{c_o}{Q} + \sum\limits_{i = 1}^{N} z_{ik}, \hspace{2mm} d = \frac{d_o (Q-1)}{Q} +N - \sum\limits_{i = 1}^{N} z_{ik}.
\end{align*}

\noindent{\bf Sample $\lambda_s$:} from $\text{Gam}\left(\lambda_s | e, f \right)$, where
 \begin{align*}
 e =\frac{NK}{2} + e_o , \hspace{2mm} f = \frac{1}{2}\sum\limits_{i=1}^{N} ||{\bf s}_i - \boldsymbol\mu_{s_{i}}||_2^2+ f_o.
 \end{align*}

\noindent{\bf Sample $\lambda_{\epsilon}$}: from $\text{Gam}\left(\lambda_{\epsilon} | g, h\right)$, where
\begin{align*}
g =  \frac{NL}{2} + g_o , \hspace{2mm} h =  \frac{1}{2}\sum\limits_{i=1}^{N} ||{\bf y}^h_i - \boldsymbol\Phi({\bf z}_i \odot {\bf s}_i)||_2^2+ h_o.
\end{align*}

\subsubsection{Prior distribution parameter computation}
\label{sec:prior}
The Gibbs sampling process estimates the posterior distributions for our model parameters. 
However, to start sampling, we require pre-computed Means $\boldsymbol\mu_{k_o}, \forall k$ and $\mu_{s_{iko}}, \forall i, k$ of the prior distributions over the basis vectors $\boldsymbol\varphi_k$ and the weights $s_{ik}$ of the representation coefficients.
Note that, the  values of the `precision' of the  prior distributions  is  not important in our approach because of the used non-informative Gamma hyper-priors.
However, the values of the Mean parameters of the prior distributions directly affect  the posterior distributions. 
We exploit this fact to encourage non-negativity  in the posterior distributions over $\boldsymbol\varphi_k$ and $s_{ik}$ by restricting the Mean parameters of the prior distributions over these parameters to be strictly non-negative values, computed by solving the following constrained sparse optimization problem:
\begin{align}
<{\bf D}_o, {\bf A}_o > = \min_{{\bf D}, {\bf A}} || {\bf Y}^h - {\bf D} {\bf A}||_F^2, \nonumber \\ \text{s.t.}\hspace{1mm} \forall i, k~ ||{\bf a}_i||_1 \leq \delta;  {\bf d}_k, {\bf a}_i \geq 0,
\label{eq:sparse}
\end{align}
where $||.||_F$ and $||.||_1$ are the Frobenius and $\ell_1$ norms respectively, ${\bf Y}^h \in \mathbb R^{L \times K}$ comprises the hyperspectral pixels (i.e. spectra) arranged as its columns,  ${\bf d}_k$ and ${\bf a}_i$ denote the $k^{\text{th}}$ and the $i^{\text{th}}$ columns of ${\bf D} \in \mathbb R^{L \times K}$ and ${\bf A} \in \mathbb R^{K \times N}$ respectively, and $\delta$ is a pre-defined constant. 

\begin{figure*}[t] 
   \centering   \includegraphics[width=\textwidth]{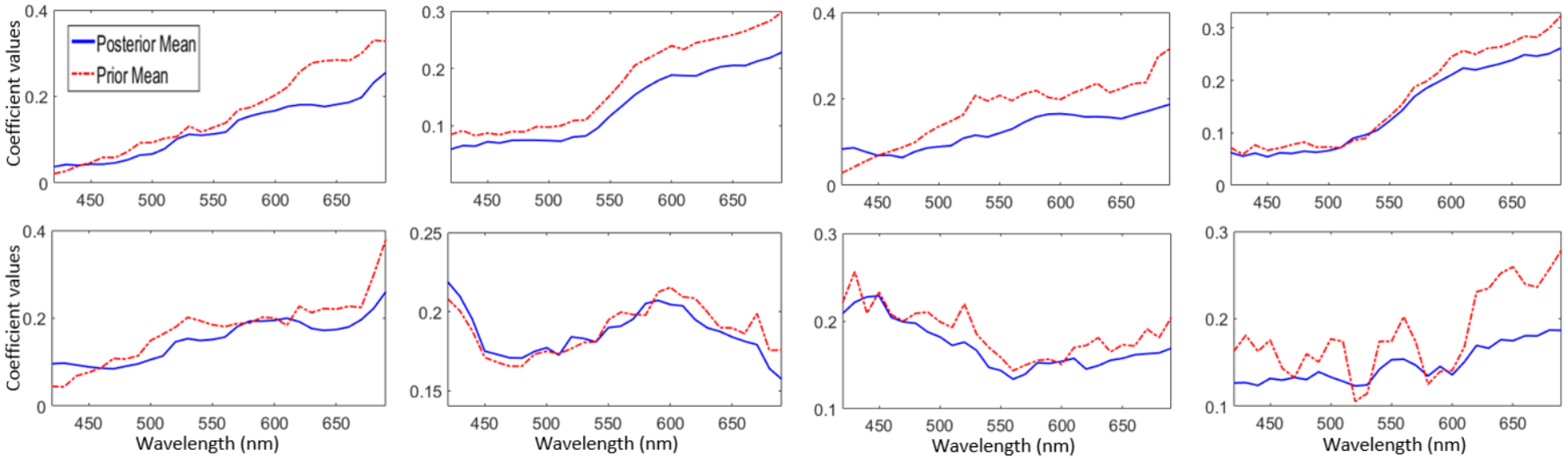} 
   \caption{{Top-left to bottom-right: The Prior Means (columns of ${\bf D}_o$) arranged in descending order of usage in data factorization. The corresponding Posterior Mean computed for each spectra is also given. The Posterior Means are significantly different from the Prior Means, and are  relatively smoother function of  wavelength.} }
   \label{fig:spectr}
\end{figure*}

We use the  columns of ${\bf D}_o$ computed in Eq.~(\ref{eq:sparse}) as $\boldsymbol\mu_{k_o}$ in our approach whereas the coefficients of ${\bf A}_o$ are used as the values of $\mu_{s_{iko}}$. Notice that,  the coefficient matrix ${\bf A}_o$ is constrained to be sparse. Furthermore,  ${\bf D}_o$ is allowed to be an over-complete dictionary for the training data. 
These properties are in-line with our representation model~(\ref{eq:mod}).
Therefore, the factorization of the training data under Eq.~(\ref{eq:sparse}) serves well for computing the prior distribution parameters for the proposed model. 
It is worth mentioning that the computed prior distribution Means only provide an informed guess for the posterior distributions in our model.
The Means of the posterior distributions estimated by our approach are generally significantly different from the values computed by solving~(\ref{eq:sparse}).
To illustrate, Fig.~\ref{fig:spectr} plots {eight} representative prior Means and their corresponding posterior values computed by the proposed approach (for CAVE database~\cite{CAVE_0293}). Details of the experiment are provided in Section~\ref{sec:Exp}.
{
From the top-left to bottom-right of the figure, the columns of ${\bf D}_o$ (i.e.~Prior Means) are arranged in descending order of their usage in data factorization under Eq.~(\ref{eq:sparse}).
Along a visible difference between the prior and the corresponding  posterior values, we can also observe that the posterior Means are smoother function of wavelength in the figure.  
This is true even for the cases where the prior spectra are already relatively smooth functions of wavelength. 
Note that, the smoother spectra in ${\bf D}_o$ generally correspond to the lower frequency components in our data factorization.  
Another important observation in the figure is that the posterior Means are completely confined to the non-negative values. }

\subsection{Testing}
As indicated in Fig.~\ref{fig:Schema}, the testing phase uses the spectrally transformed (posterior) Means of the Gaussian Processes to represent the ($p \times p \times 3$-dimensional) patches of the test RGB image. The transformed Means are stored in `$C$' matrices $\widetilde{\boldsymbol\Phi}_c, c \in \{1, 2,...,C\}$.
To choose an appropriate $\widetilde{\boldsymbol\Phi}_c$ to represent a given RGB patch, the latter is first assigned a cluster of similar RGB patches by computing its Euclidean distances from the cluster centroids. 
These clusters are directly computed by spectrally transforming the already clustered hyperspectral patches, under the same transformation that is applied to the Mean parameters of the Gaussian Processes. 
The transformation of the clustered hyperspectral patches is an off-line process in our approach and it is performed only once for a given training dataset. 

Once the RGB patch is assigned its cluster, the corresponding $\widetilde{\boldsymbol\Phi}_c$ is used to represent the patch by solving  the following optimization problem for each pixel ${\bf y}_i, i\in \{1,2...,p^2\} $:
\begin{align}
<\boldsymbol\beta_i> = \min_{\beta_i}||\boldsymbol\beta_i||_1, \hspace{2mm} \text{s.t.}~ ||{\bf y}_i - \widetilde{\boldsymbol\Phi}_c\boldsymbol\beta_i||_2^2 \leq \delta_1,
\label{eq:Lasso}
\end{align}
where $\delta_1$ is a pre-defined constant. The target hyperspectral pixel ${\bf y}^h_i$ is computed by taking the linear combination of the Gaussian Process Means in $\mathcal G_c$ weighted by the coefficients of $\boldsymbol\beta_i$.
In Eq.~(\ref{eq:Lasso}), we  constrain  $\boldsymbol\beta_i$ to be sparse by restricting its $\ell_1$-norm, which is to match the underlying sparisty of the spectral signals captured by our representation model with the Beta-Bernoulli distributions.
Our approach extracts overlapping patches from the RGB image and performs the above mentioned process for each patch. A pixel in the estimated hyperspectral image is computed as the Expected value of the reconstructed spectra for that pixel.

\section{Implementation details}
\label{sec:ImpD}
We implement our approach using Matlab, running on an Intel Core i7 CPU at 3.6~GHz with 32~GB RAM.
To solve the optimization problems~(\ref{eq:sparse}) and (\ref{eq:Lasso}), we used the SPAMS library~\cite{SPAMS} that implements  the Online Dictionary Learning method~\cite{mairal2009online} to solve~(\ref{eq:sparse}) and the non-negative variant of the LARS algorithm~\cite{lars} to solve~(\ref{eq:Lasso}). 
The values of the parameters $\delta$ and $\delta_1$ for these problems are  empirically chose to be 0.01 and $10^{-21}$, respectively.
It is worth mentioning that the precise value of $\delta$ is relatively unimportant for our approach as it only influences the prior parameters, for which posterior values are again computed. 
{However, a small value of $\delta_1$ is preferable for our method that aims to compute an accurate mapping of a given RGB image to a hyperspectral image. We note that the SNR of RGB image can influence this parameter value. A low SNR of RGB image should be accounted for in Eq.~(\ref{eq:Lasso}) by choosing a larger reconstruction error $\delta_1$. The value chosen in our experiments results in the best performance under the used evaluation protocol.}

For our Bayesian model (\ref{eq:mod}), we chose the values of the parameters based on the parameter significance. 
Except $c_o$ and $d_o$, the parameters $a_o$ to $h_o$ belong to the non-informative Gamma hyper-priors in our model.
Therefore, the inferred posterior distributions largely remain insensitive to their small values, e.g. in the range $[10^{-3}, 10^{-9}]$. This fact is easily verifiable from the sampling expressions in Section~\ref{sec:Infer}. We chose $10^{-6}$ as the value for these parameters.  
Following \cite{Akhtar:2015}, we used the same values for $c_o$ and $d_o$, i.e. $10^{-6}$. 
For the Beta distributions, we set the value of $Q$ equal to  the number of Gaussian Processes to be learned, i.e. $K$, whereas $K$ itself is automatically decided by the cluster size.
The ratio of $K$ to the total number of Gaussian Processes to be learned for the complete training data is kept the same as the ratio of the number of patches in the cluster to the total number of training patches.

Based on the fact that our training data is largely clean in terms of Gaussian noise, we set the precision of the  noise $\lambda_{\epsilon_o} = 10^6$.
Similarly, $\lambda_{s_o}$ was also set to the same value because we set the initial values of $s_{ik}$ equal to the computed $\mu_{s_{iko}}$, implying that we believe  $\mu_{s_{iko}}$ to be already a reasonable estimate of the posterior distribution  over $s_{ik}$.
The parameters $z_{ik}$ were simply initialized to the support of $s_{ik}$.
We also initialized $\boldsymbol\Phi$ with the values of  the prior Mean parameters of the Gaussian Processes.  
All the Bernoulli parameters $\pi_{k_o}$ were initialized to the non-informative value $0.5$. For the Gaussian Process Kernels, we chose $\ell = 3$ as $1/10^{\text{th}}$ of the total number of channels in the resulting hyperspectral image.

Note that, due to the Bayesian nature of our approach, meaningful interpretations can be associated with most of the model parameters.
We used these interpretations to guide us in choosing the initial values of the parameters. 
Nevertheless, we also tested the approach with multiple other close-by initial parameter values and found that the results remain fairly robust to those different initializations, which is typical for a Bayesian approach.

\section{Experimental evaluation}
\label{sec:Exp}
We evaluated the proposed approach using the hyperspectral images from CAVE database~\cite{CAVE_0293},  iCVL database~\cite{Arad2016} and  Harvard database~\cite{chakrabarti2011statistics}.
To benchmark, the approach is compared with the related  technique of hyperspectral recovery from the RGB images proposed by Arad and Ben-Shahar~\cite{Arad2016}, {and  Aeschbacher et al.~\cite{Aeschbacher}, i.e. A+}. 
We carefully implemented~{\cite{Arad2016}} (public code is not available) using the SPAMS library~\cite{SPAMS} and the efficient implementation of the K-SVD algorithm~\cite{ksvd} provided by Rubinstein et al.~\cite{ksvd1}, {and used the author provided code for A+~\cite{Aeschbacher}}.
In the original work~\cite{Arad2016}, the authors proposed to use the Orthogonal Matching Pursuit (OMP) algorithm~\cite{omp} to perform sparse coding of the test pixels over a known dictionary. 
Our experiments showed that sparse coding using the LARS algorithm~\cite{lars}  instead, consistently improves the accuracy of their method\footnote{This observation is also well supported by the existing literature~\cite{myTGRS1}, \cite{5692827}. The greedy pursuit strategy of OMP is more susceptible to the commonly found high mutual coherence of the spectral dictionaries as compared to the LARS algorithm~\cite{lars}.}. Therefore, we performed the sparse coding by solving~(\ref{eq:Lasso}) in our implementation. 
This also allowed for a more fair comparison as we are able to use the same value of $\delta_1$ (i.e. $10^{-21}$) for~\cite{Arad2016} and our approach. 
For the CAVE and the iCVL databases we also provide the results of employing our framework in a non-Bayesian settings. That is, instead of computing the Gaussian Processes, we directly use the bases computed by the Online Dictionary Learning method~\cite{mairal2009online} in-place of the Gaussian Process Mean parameters in our framework.
We denote this variant of our technique as `DL' (for Dictionary Learning). The approach in~\cite{Arad2016} is mentioned as `Arad' (after the author's name).
For each experiment, the total number of the Gaussian Processes in our approach are kept exactly the same as the total number of the dictionary atoms used for {the other approaches}, i.e. $1,000$ for the CAVE and Harvard images and $2,000$ for the iCVL images. These parameters were optimized for~\cite{Arad2016} using the cross validation technique and the values were directly used {for the other approaches}.
In all the experiments, we fixed the number of clusters `$C$' to 10, and used $8 \times 8$ spatial patches. 

In our experiments, we treated the hyperspectral image from a database as the ground truth  and transformed it to the RGB image using the spectral response of the Nikon D700 camera. 
We adopt this protocol from the RGB+hyperspectral image fusion literature~\cite{Akhtar:2014a}, \cite{Kawakami:2011:HHI:2191740.2191889}, \cite{Lanaras:2015}.
We used the Root Mean Square Error (RMSE) and the Spectral Angle Mapper (SAM)~\cite{yuhas1992discrimination} as the evaluation metrics.  
Whereas RMSE is  useful in analyzing the reconstruction of the individual spectral channels of the estimated hyperspectral image, SAM provides a better assessment of the image reconstruction  along the spectral dimension. 
The SAM value indicates the angle between the computed and the ground truth  spectra (in degrees) - a lower value is more desirable. 
In the case of iCVL database, many images contain a large contrast in the scene  illumination due to the presence of shadows in the images. Therefore, to avoid bias in the RMSE values we used the relative RMSE for iCVL images, computed by normalizing the error with the ground truth.

\subsection{CAVE images}
\label{sec:CAVE}
The CAVE database~\cite{CAVE_0293} consists of hyperspectral images of 32 daily life object scenes, acquired by the cooled CCD camera Apogee Alta U260 in conjunction with the Varispec liquid crystal tunable filter under the CIE standard  illuminant D65.
The $512\times 512 \times 31$-dimensional hyperspectral images are collected in the wavelength range 400nm - 700nm, with 10nm steps. 
The imaged scenes contain a variety of objects with distinct spectral responses, which makes the hyperspectral recovery for the CAVE images particularly challenging.
Due to low signal-to-noise ratio and blur, we removed the first two and the last spectral channels of the hyperspectral images of this dataset in our experiments. 
We also do the same for the iCLV and the Harvard images in the Sections to follow. 

\begin{figure*}[t] 
   \centering   
   \includegraphics[width=5.3in]{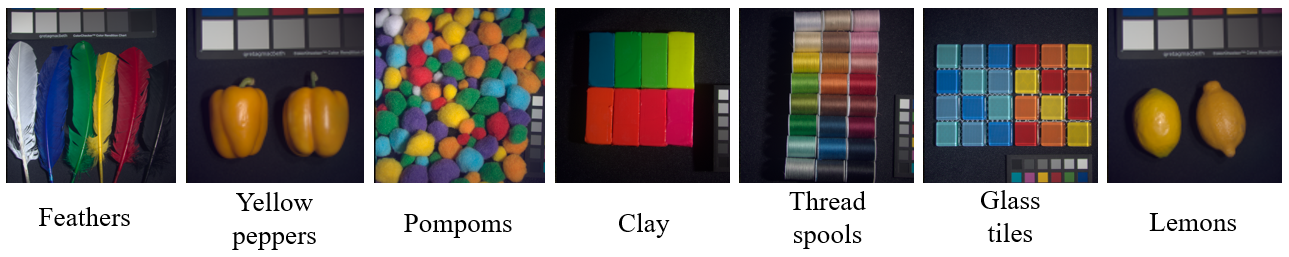} 
   \caption{RGB images from the CAVE database~\cite{CAVE_0293} used in Table~\ref{tab:CAVE}. The variety of colors makes the images more challenging for spectral recovery.} 
   \label{fig:CAVE_sample}
\end{figure*}

For evaluation, we randomly selected seven different scenes  from the CAVE  dataset for testing.
The RGB images of the testing scenes are shown in Fig.~\ref{fig:CAVE_sample}.  The variety of the spectral sources is easily noticeable in the chosen scenes by considering the different colors present in the images.
To learn the Gaussian Processes, we randomly selected five different hyperspectral  images from the database for each test scene, ensuring that the test scene is not present among those images. 
The same training images are used for all approaches in our experiments.
Table~\ref{tab:CAVE} summarizes the results of experiments, with best results bold-faced.  

\begin{table}
\label{tab:CAVE}
\caption{Results on CAVE images~\cite{CAVE_0293}: The RMSE values are computed in the range of 8-bit images and SAM is measured in degrees. The DL-variant of our method does not exploit Gaussian Processes.}
\tabcolsep=0.05cm
\centering
\begin{tabular}{cc cccc cccc c}
\hline
\multicolumn{1}{c@{\quad}}{{\bf Images}} 
&& 
\multicolumn{4}{c}{ {\bf RMSE}}    
&&                                           
\multicolumn{4}{c}{{\bf SAM}} 
\\ 
\cline{1-1} \cline{3-6}\cline{8-11}
&& Prop. & DL & Arad~\cite{Arad2016}  & {A+~\cite{Aeschbacher}}     && Prop. & DL & Arad~\cite{Arad2016} & { A+~\cite{Aeschbacher}}  \\ \cline{3-6}\cline{8-11}
Feathers			&& {\bf 7.89} 			& 8.11 	& 9.17  	& {8.56}   		&& {\bf 12.02} 	& 13.71 	& 15.06 & {13.09}  \\ 
Peppers 			&& {\bf 5.74}			& 7.18 	& 7.64  	& {6.99}    	&& {\bf 16.34} 	& 17.76 	& 18.45 & {17.64}  \\ 
Pompoms 			&& {\bf 7.28} 			& 7.84 	& 9.08 		& {8.13}    	&& {\bf 9.09} 	& 9.78 		& 10.42 & {9.86}  \\ 
Clay 				&& {\bf 8.01} 			& 9.07 	& 8.44  	& {8.86}    	&& {\bf 14.82} 	& 15.91 	& 16.78 & {16.01}  \\ 
Spools 				&& {\bf 6.22} 			& 6.76 	& 7.01  	& {6.57}    	&& {\bf 16.15} 	& 16.77 	& 18.88 & {17.13}  \\ 
Glass tiles			&& {\bf 6.68}			& 7.10  & 7.71		& {7.36}   		&& {\bf 11.71} 	& 12.48 	& 13.37 & {11.76}  \\
Lemons				&& {\bf 3.42}  			& 4.01 	& 5.03  	& {4.93}    	&& {\bf 18.14} 	& 19.13 	& 22.99 & {18.83}  \\
 \hline
\end{tabular}
\end{table}

\begin{figure*}[t] 
   \centering   
   \includegraphics[width=5.2in]{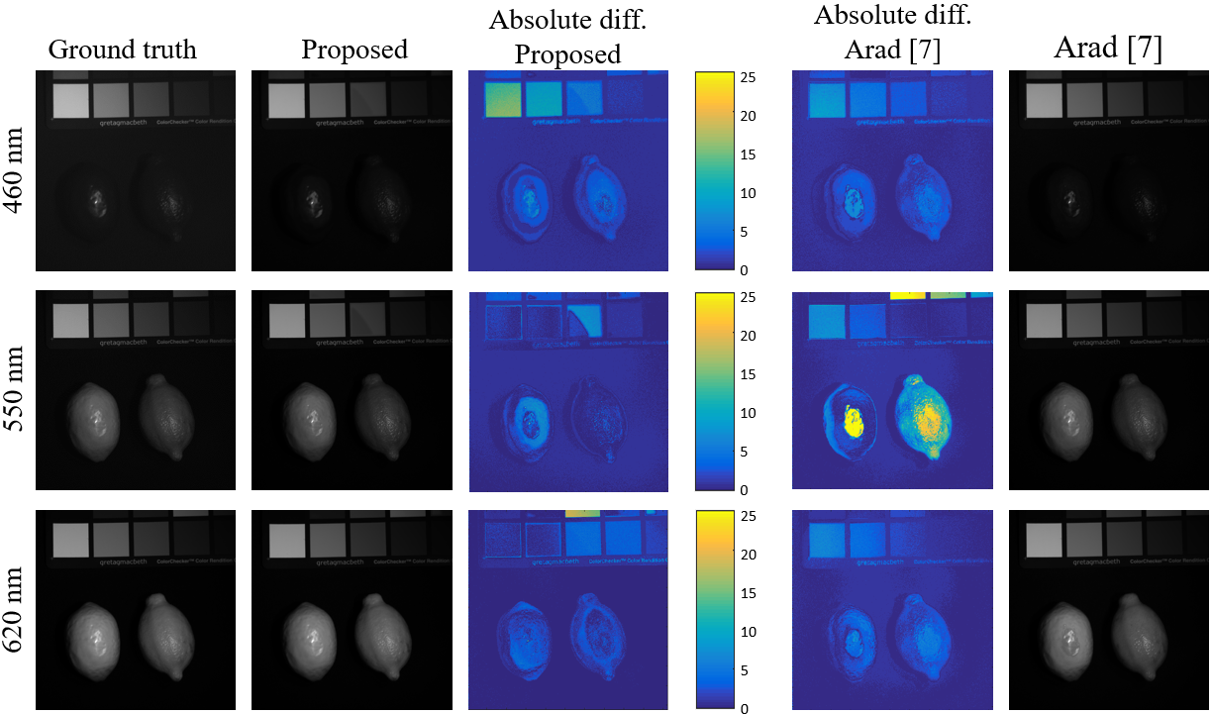} 
   \caption{Recovery of the spectral images of `Lemons' scene from CAVE~\cite{CAVE_0293} database: Along the ground truth at wavelengths 460, 550 and 620nm, recovered spectral images by the proposed approach and Arad~\cite{Arad2016} are shown. The absolute differences between the recovered images and the ground truth are also given in the range of 8-bit images.} 
   \label{fig:CAVE_res}
\end{figure*}

From the table, we can see that the proposed approach is able to consistently achieve good performance for all the test images. Not surprisingly, the DL-variant of our framework also achieves a reasonable performance. However, it does not perform as well. 
It is worth emphasizing that the DL-variant uses all the steps in our framework just as the proposed approach, except that it does not take advantage from the Gaussian Processes. This demonstrates the usefulness of Gaussian Processes in spectral recovery. 
In comparison to Arad~\cite{Arad2016}, the proposed  approach achieves an average $16.35 \%$ reduction in the RMSE values of the test images and $9.39\%$ reduction   in the average SAM value. 
{The corresponding improvements over  A+~\cite{Aeschbacher} are $11.98\%$, and $5.85\%$ respectively.}
In Fig.~\ref{fig:CAVE_res}, we also illustrate the recovery of the spectral channels of the `Lemons' scene in the CAVE database.  
The figure provides recovered intensity maps for the spectral channels corresponding to the wavelengths  460, 550 and 620 nm for our approach and for Arad~\cite{Arad2016}, along the ground truth. The absolute differences of the recovered images with the ground truth are also shown in the range of 8-bit images.
{We prefer showing the spectral bands for Arad~\cite{Arad2016} as reference due to the similarity of that technique with the proposed approach.}

\begin{figure*}[t] 
   \centering   
   \includegraphics[width=5.2in, height = 1in]{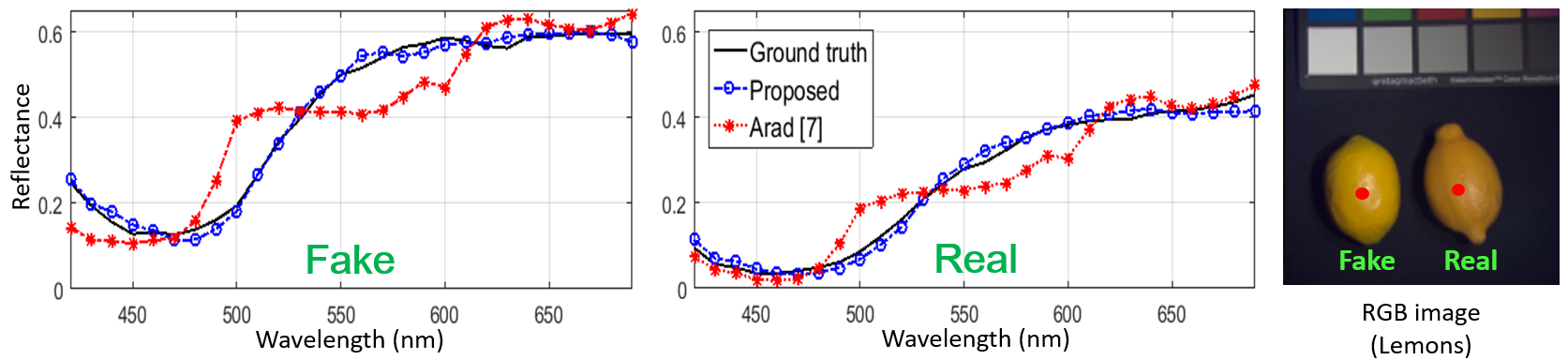} 
   \caption{Illustration of spectral recovery for `Lemons' scene from the CAVE~\cite{CAVE_0293} database: The spectra are sampled from a \emph{single pixel} on each Lemon from the area spotted `red' in the RGB image on the right.  } 
   \label{fig:CAVE_lemon}
\end{figure*}

Interestingly, the Lemon scene contains a `real' and a `fake' Lemon that are not easily distinguishable by observing only the RGB image of the scene.
One can hope to solve this problem by hyperspectral imaging. 
In Fig.~\ref{fig:CAVE_lemon}, we show the spectral profiles of single pixels (taken from the areas spotted red in the RGB image on the right) for the two Lemons. 
The spectral profiles are provided for the ground truth, the proposed approach and  Arad~\cite{Arad2016}. 
We can see that our approach recovers the spectral profiles accurate enough to  distinguish between the Lemons almost as reliably as they can be differentiated using the ground truth. 
{Although the shown ground truth spectra of real and fake Lemons  appear qualitatively similar, the SAM value for these spectra is 5.5, whereas the RMSE value is 19.2 (in 8-bit format). The average SAM for the spectra reconstructed by our approach and the ground truth for the opposite type of Lemon (i.e.~fake for real and vice versa) is 5.1, which is close to 5.5. Correspondingly, the average RMSE value for our approach is 18.9, demonstrating that the reconstructed spectra provide a well founded  clue for distinguishing between the real and fake Lemons. The average SAM and RMSE values between the reconstructed spectra and the ground truth of the correct type of Lemon are 0.9 and 2.7, respectively.}
This analysis also illustrates the practical usefulness of estimating spectral details from RGB images. 
The relative smoothness of the profiles recovered by the proposed approach can be directly attributed to the use of Gaussian Process in modeling the natural spectra.

\subsection{iCVL images}
\label{sec:icvl}

The iCVL dataset~\cite{Arad2016} consists of 100 hyperspectral images captured by Specim PS Kappa DX4 camera with a rotary stage for the spatial scanning. The provided images are $1392 \times 1300 \times 31$ dimensional cubes, with $31$ spectral channels in the range 400 - 700 nm, with 10 nm steps\footnote{The camera captures 519 spectral bands in the range 400 - 1000 nm, with 1.25 nm increments. Arad and Ben-Shahar~\cite{Arad2016} computationally reduced the images to 31 bands to match the existing  datasets.}. The images pertain to a variety of indoor and outdoor scenes that can be divided into different domains, e.g. Park, Beach, Green house etc. We used this division to evaluate the domain-specific and cross-domain performance of our approach.


\begin{table}
\caption{{Domain-specific results on iCVL dataset~\cite{Arad2016}: Average relative Root Mean Square Error (RMSE) ($\times 10^{-2}$) and Spectral Angle Mapper (SAM) are reported for three images from each domain.}}
\tabcolsep=0.05cm
\centering
\begin{tabular}{cc cccc cccc c}
\hline
\multicolumn{1}{c@{\quad}}{{\bf Images}} 
&& 
\multicolumn{4}{c}{ {\bf RMSE}}    
&&                                           
\multicolumn{4}{c}{{\bf SAM}} 
\\ 
\cline{1-1} \cline{3-6}\cline{8-11}
&& Prop. & DL & Arad~\cite{Arad2016}  & {A+~\cite{Aeschbacher}}     && Prop. & DL & Arad~\cite{Arad2016} & { A+~\cite{Aeschbacher}}  \\ \cline{3-6}\cline{8-11}
Park 				&& {{\bf 8.49}} & {9.75}  & {11.19} & {8.99}  && {{\bf 5.13}} & {5.69} & {6.23} & {5.79}\\ 
Green house 		&& {{\bf 7.61}} & {8.83}  & {9.41}  &	{9.06}  && {{\bf 3.89}} & {4.72} & {5.63} & {4.81} \\ 
Outdoor 			&& {{\bf 6.99}} & {8.31}  & {8.72}  &	{ 8.54}  && {{\bf 3.68}} & {4.00} & {4.59} & {4.59}\\
Floral 				&& {{\bf 7.38}} & {7.95}  & {8.08}  &	{ 7.77}  && {{\bf 4.49}} & {4.94} & {5.22} & {4.87} \\ 
Beach 				&& {{\bf 4.33}} & {4.89}  & {5.21}  &	{5.09}  && {{\bf 2.31}} & {3.05} & {3.55} & {3.11} \\  
Objects 			&& {{\bf 1.51}} & {1.97}  & {2.05}  &	{2.00}  && {{\bf 0.82}} & {1.10} & {1.29} & { 1.20} \\ \hline
\end{tabular}
\label{tab:ICLV}
\end{table}

\begin{figure*}[t] 
   \centering
   \includegraphics[width=5.3in]{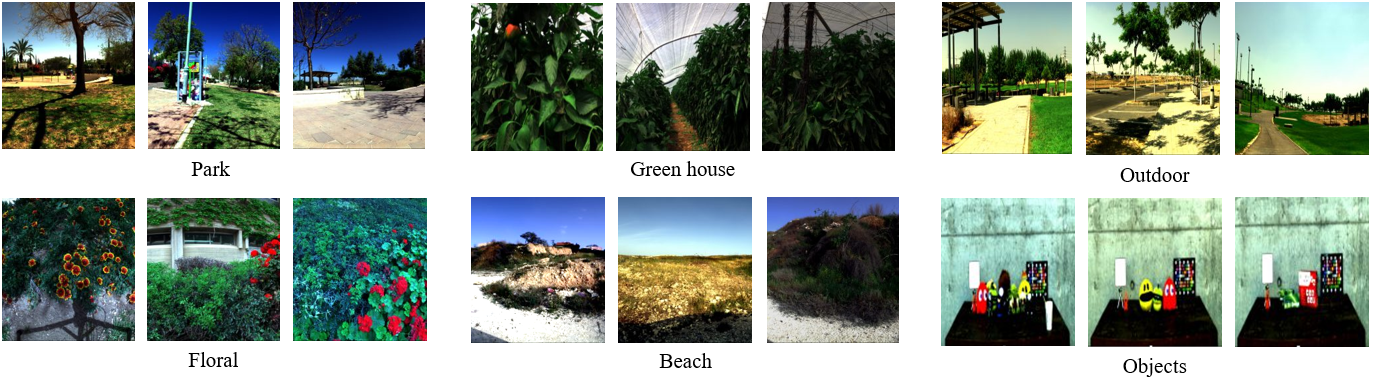} 
   \caption{ {RGB images of representative scenes from the six domains in iCVL database~\cite{Arad2016} used in Table~\ref{tab:ICLV}.} }
   \label{fig:ICLV_sample}
\end{figure*}

\begin{figure*}[t] 
   \centering   
   \includegraphics[width=5.2in]{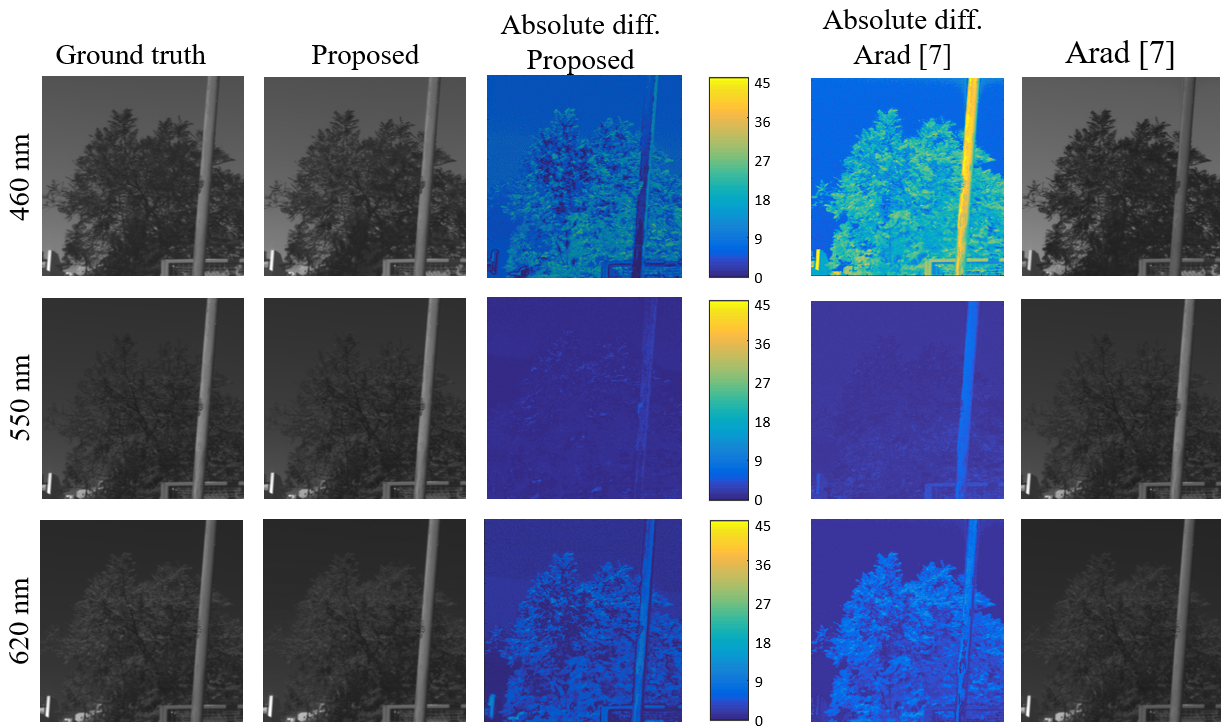} 
   \caption{Example spectral channel recovery of $512 \times 512$ dimensional spatial patch from a scene in the Park domain from iCVL dataset~\cite{Arad2016}. Along the ground truth at wavelengths 460, 550 and 620nm, recovered spectral images and the absolute differences between the recovered images and the ground truth are also given in the range of 8-bit images.} 
   \label{fig:ICLV_res}
\end{figure*}

For the domain-specific experiments, we selected a set of five scenes per domain (as allowed by the dataset) and reconstructed the hyperspectral images {for each scene such that the} remaining  scenes we used in the training. We report the average relative RMSE and SAM values for our reconstructions in Table~\ref{tab:ICLV}.
Again, the proposed approach consistently performed well on all the tested scenes,  
improving the average relative RMSE over Arad~\cite{Arad2016} by {$18.62\%$} and the average SAM value by  {$23.35\%$}.
{The improvements over A+~\cite{Aeschbacher} are $12.31\%$ and $16.78\%$, respectively.}
Considering the overall low SAM and RMSE values, we can claim that domain-specific spectral recovery by our technique is reasonably accurate. 
Naturally, the results improve with the similarity of the spectral sources in the scenes for training and testing.
In Fig.~\ref{fig:ICLV_sample}, we show the RGB images of the three {representative} scenes per domain used in our experiments. 
In Fig.~\ref{fig:ICLV_res}, we also provide the recovered spectral channels at 460, 550 and 620~nm for one of the scenes from the Park domain.
For a better visualization, the spectral images are provided for a $512 \times 512$ spatial patch. 

{In this work, we propose to use training data from the same/similar scene domains as that of the test image. That is, for appropriate spectral recovery we prefer Gaussian Processes to model spectral sources of similar nature to what we can expect in the test scene. We do not claim that the proposed method can precisely  recover spectra of e.g.~grass using only the training spectra of e.g.~clouds. Nevertheless, the Gaussian Processes that model a variety of spectral sources can still be expected to provide an acceptable estimate of the spectral details for different objects. To demonstrate  that, we report the results of recovering hyperspectral images of the scenes from Floral domain in Fig.~\ref{fig:ICLV_sample} using the training data from (1) Outdoor domain and  (2) Green house domain.
The results of these experiments are reported in Table~$3$. Although there is some  degradation in performance as compared to Table~$2$, the results remain acceptable despite a significant contrast of spectral sources in the scene domains.  }


{ We also performed exhaustive experiments with the iCVL database following the protocol of \cite{Arad2016}. For that, we reconstructed the hyperspectral image of each scene in the database by using all the remaining scenes as the training data. The average relative RMSE and SAM values for our approach for those experiments are $0.0601$ and $2.49$, respectively. The respective values for Arad~\cite{Arad2016} in our experiments were  $0.0814$ and $3.64$, and the corresponding values for A+~\cite{Aeschbacher} were $0.0673$ and $2.95$, respectively.}

\begin{table}
\caption{{Cross-domain results: Average relative RMSE ($\times 10^{-2}$) and SAM value of the approaches when each image from one domain is constructed using the training data from the other domain.} }
\tabcolsep=0.09cm
\centering
\begin{tabular}{cc cc cc c}
\hline
\multicolumn{1}{c@{\quad}}{{\bf Domain}} 
&& 
\multicolumn{2}{c}{ {\bf relative RMSE}}    
&&                                           
\multicolumn{2}{c}{{\bf SAM}} 
\\ 
\cline{1-1} \cline{3-4}\cline{6-7}
&& Prop. & Arad~\cite{Arad2016}       && Prop. & Arad~\cite{Arad2016}   \\ \cline{3-4}\cline{6-7}
Floral from Outdoor 			&& {{\bf 9.44}} & { 11.53}     && {{\bf 5.28}} & {6.77} \\
Floral from Green house 		&& {{\bf 10.06}} & {10.39}     && {{\bf 5.05}} & {5.92} \\
\hline 
\end{tabular}
\label{tab:ICLV2}
\end{table}

\subsection{Harvard images}
\label{sec:Harvard}
\begin{figure*}[t] 
   \centering   
   \includegraphics[width=5.7in]{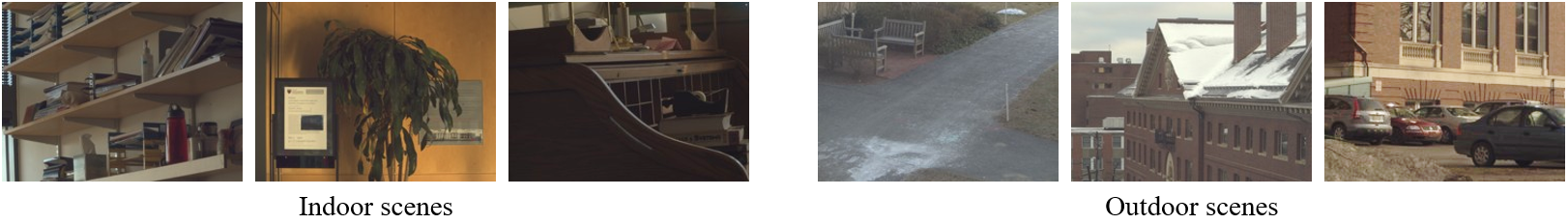} 
   \caption{RGB images of the Harvard scenes~\cite{chakrabarti2011statistics} used for testing in Table~\ref{tab:Harvard}. The domain labels are also mentioned.} 
   \label{fig:Harvard_sample}
\end{figure*}

\begin{figure*}[t] 
   \centering   
   \includegraphics[width=5.2in]{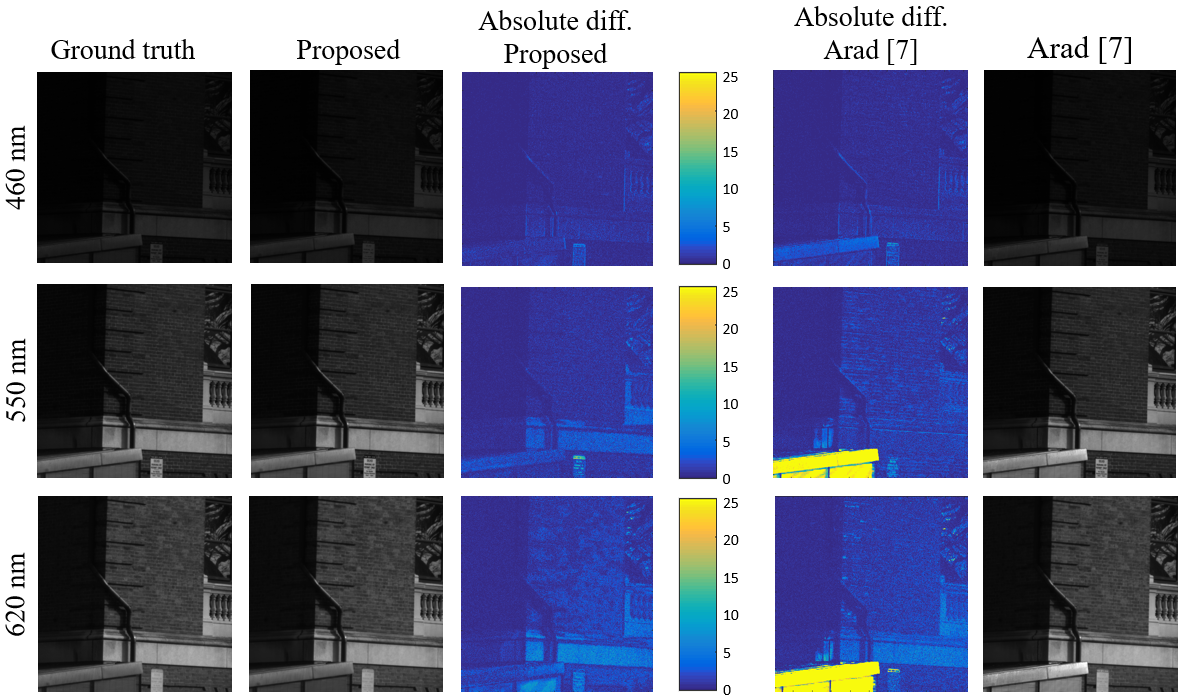} 
   \caption{Example spectral recovery of $512 \times 512$ spatial patch of an Outdoor scene from the Harvard dataset~\cite{chakrabarti2011statistics}. The results are given in the range of 8-bit images.} 
   \label{fig:Harvard_res}
\end{figure*}

The real-world hyperspectral images in the Harvard database~\cite{chakrabarti2011statistics} are acquired in the daylight illumination using Nuance FX CRI Inc.~camera that uses an integrated liquid crystal tunable filter.  The $1040\times1392\times 31$-dimensional image cubes sample the scene radiance in the spectral wavelength range 420 - 720~nm with a 10~nm interval. The dataset provides hyperspectral images of a variety of Indoor and Outdoor scenes.  To evaluate the spectral recovery of the Harvard scenes, we randomly chose a subset of eight images each from the Indoor and the Outdoor domains. 
From these subsets, we reconstructed the hyperspectral images of three random scenes by using the remaining  seven scenes for each image in training. 
The average RMSE and SAM values for the test scenes for both domains are reported in Table~\ref{tab:Harvard}. 
The RGB images of the test scenes are provided in Fig.~\ref{fig:Harvard_sample}. 
Over Arad~\cite{Arad2016}, the average RMSE improvement for our approach is $30.75\%$ and the average SAM improvement is $9.85\%$.
We also provide an example of the spectral channel reconstruction by our approach and Arad~\cite{Arad2016} in Fig.~\ref{fig:Harvard_res}.

\begin{table}
\caption{Results on Harvard images~\cite{chakrabarti2011statistics}: The RMSE values are computed in the range of 8-bit images and SAM is measured in degrees. }
\tabcolsep=0.06cm
\centering
\begin{tabular}{cc ccc ccc c}
\hline
\multicolumn{1}{c@{\quad}}{{\bf Domain}} 
&& 
\multicolumn{3}{c}{ {\bf RMSE}}    
&&                                           
\multicolumn{3}{c}{{\bf SAM}} 
\\ 
\cline{1-1} \cline{3-5}\cline{7-9}
&& Prop. & Arad~\cite{Arad2016} &{A+~\cite{Aeschbacher}}       && Prop. & Arad~\cite{Arad2016}  & {A+~\cite{Aeschbacher}}  \\ \cline{3-5}\cline{7-9}
Outdoor scenes 		&& {\bf 4.36} & 6.81 & {5.32}    && {\bf 3.67} &4.62 & {3.99}\\
Indoor scenes 		&& {\bf 2.78} & 3.50 & {3.11}   && {\bf 7.13} & 7.36 & {7.29}\\
\hline 
\end{tabular}
\label{tab:Harvard}
\end{table}

{
\subsection{Real-world experiments}
We also performed real-world experiments where we reconstructed the reflectance spectra of a Macbeth color chart using learned Gaussian Processes. We acquired color images of the Macbeth chart using an available Canon 750D DSLR camera, and  reconstructed its reflectance with the help of Gaussian Processes learned from the entire CAVE database~\cite{CAVE_0293}, using the same parameter settings as in Section~\ref{sec:CAVE}. As noted in Section~\ref{sec:PF}, our approach expects the spectral transformation between the target hyperspectral signatures and the acquired color image as an input. It is possible to  precisely compute this transformation using the hyperspectral sensor employed  to collect the training data and the color camera~\cite{Lanaras:2015}. However, since we use a public hyperspectral database in our experiments, we approximate the desired transformation ${\bf T} \in \mathbb R^{3 \times 28}$ as: ${\bf T} = (({\bf Y}^{h~\intercal})^{\dagger} {\bf Y}^{\intercal})^{\intercal}$, where $\dagger$ denotes a Moore-Penrose inverse of the matrix, ${\bf Y}\in \mathbb{R}^{3 \times n}$ contains $n$ RGB pixels, and ${\bf Y}^h\in \mathbb{R}^{28 \times n}$ are the spectral samples from the CAVE database for the same spectral sources.  
In our experiments, the computed matrix ${\bf T}$ only approximates the spectral transformation between the acquired color images and the CAVE database, however our results demonstrate that this already leads to a useful recovery of the spectral details.  

In Fig.~\ref{fig:Macbeth}, we show the spectra reconstructed by our approach for seven random swatches of the Macbeth chart along the ground truth spectra. The ground truth is computed with the help of a StellarNet CXR-SR-50 spectrometer that measures spectral signals in the wavelength range 224 - 1100~nm with a fixed interval of 0.5 nm. Since the Gaussian Processes are learned from the images that have 28 channels that are 10 nm apart, we re-sampled the ground truth spectra to match the channels of the training data. The ground truth and the reconstructed spectra shown in the figure are the mean values computed over ten random points per swatch.

In general, a good recovery of the spectral details are observable in Fig.~\ref{fig:Macbeth}. Each plot of the figure also shows the RMSE and SAM values for the recovered spectra. The shown spectra are chosen randomly as representative examples. The average SAM and RMSE values computed over all the swatches of the chart were 6.1 and 6.9, respectively.

\begin{figure*}[t] 
   \centering   
   \includegraphics[width=7
   in]{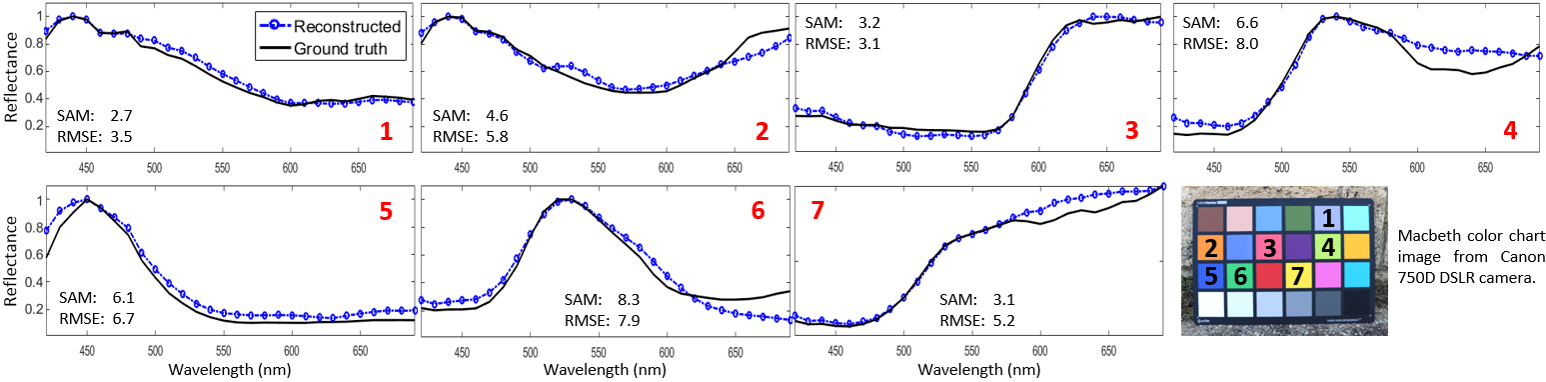} 
   \caption{{Spectral recovery of Macbeth color chart swatches using Gaussian Processes learned from CAVE database~\cite{CAVE_0293}: Reconstructed spectra of representative swatches are shown. The used RGB image is also displayed with swatch numbers indicating the plots in the figure. The shown reconstruction and the ground truth are the mean values of ten random points on a given swatch. }} 
   \label{fig:Macbeth}
\end{figure*}

}

\subsection{Computational time}
\label{sec:compute}
To train our model, we used contiguous disjoint patches of the training hyperspectral  images. 
Owing to the similarity of the patches in a cluster, we randomly sampled  $1\%$ of the pixels in each patch and used those to infer the Gaussian Processes.
Our experiments showed that due to the redundancy of the spectral signal sources in the patches, the accuracy of our approach is not significantly affected by this simplification, however it decreases the computational time by nearly two orders of magnitude. 
The average training time required for our approach for the CAVE images was computed to be 7.32 minutes.
For the Harvard and the iCVL datasets these timings were 39.63  and 54.98 minutes, respectively. 
The number of training patches for the Harvard database were $\sim 7$ times more than those for the CAVE images, whereas the iCVL data had $\sim 6$ times more patches than the CAVE images.
Moreover, $2,000$ Gaussian Processes were learned for the iCVL scenes as compared to $1,000$ Processes for the Harvard and the CAVE scenes. 
Considering these number, we can argue that the computational complexity of the overall approach remains linear in the number of training samples and the number of Gaussian Processes to be inferred.  
Our experiments demonstrated that $500$ iterations of Gibbs sampling process were always sufficient to accurately infer the desired Gaussian Processes.

For testing, the proposed approach uses overlapping patches of the RGB image to reconstruct the hyperspectral image patches with the stride of 2, and then computes the final image as their average. For the $512\times 512$ spatial patch/image reconstruction, the average timings of our approach for the CAVE, Harvard and the iCVL scenes are $3.96$, $3.81$ and $7.44$ minutes, respectively. 
{The corresponding timings for the DL method are $3.88$, $3.79$ and $7.09$ minutes.
Notice that the computational complexity during testing mainly comes from solving the sparse optimization problem in Eq.~(3), which is required by the DL method as well as Arad~\cite{Arad2016}. Since \cite{Arad2016} solves Eq.~(\ref{eq:Lasso}) once, but with a much larger dictionary, it was observed to be nearly twice as fast as the proposed approach in our experiments. However, the proposed approach leads to significant improvement in spectral recovery over~\cite{Arad2016}.}
For the complete images of the Harvard and the iCVL scenes the average timings for the proposed method are $23.35$ and $39.98$ minutes, respectively.
The above mentioned timings are for our implementation in Matlab, running on an Intel Core i7 CPU at 3.6~GHz with 32~GB RAM. 
{Since spectral recovery is widely considered an off-line process~\cite{Arad2016}, \cite{nguyen2014training}, \cite{Antonio2015}, we do not particularly optimize our implementation from the computational perspective. Nevertheless, our approach can be easily parallelized by independently  solving Eq.~(\ref{eq:Lasso}) for individual image patches.}

\section{Conclusion}
\label{sec:Conc}
In this work, we proposed a technique to computationally estimate the hyperspectral image of a scene from its RGB image of known spectral quantization. 
The proposed approach assumes a hyperspectral training prior over the imaged scene and extracts clusters of similar patches from the training images.
These clusters are used for computing multiple sets of Gaussian Processes such that each Process signifies a spectral signal source in a cluster.
The physical attributes of the spectral signatures are incorporated in the Gaussian Processes  by their  Kernels and the use of non-negative Mean prior probability distributions. 
To compute the Processes, we proposed a Bayesian representation model and  developed its inference scheme.
The computed Gaussian Processes are transformed to match the spectral quantization of the test RGB image.
To construct the target hyperspectral image from the RGB image, patches are extracted from the latter and are matched with the RGB patches created by transforming the training hyperspectral patches.
The test RGB patches are represented by the transformed Gaussian Processes for the matched clusters and the representations are subsequently combined with the original Gaussian Processes to construct the desired hyperspectral image.
We evaluated our approach by recovering 37 different hyperspectral images from the RGB images of Objects, Indoor and Outdoor scenes. In general, the results show a reasonable recovery of the detailed spectra from RGB images.
We compared the proposed approach to a recent closely related technique~\cite{Arad2016}. Our experimental results demonstrate that the proposed approach achieves up to $30.75\%$ reduction in the RMSE and $20.93\%$ reduction in the SAM values for the reconstructed images as compared to the existing approach.

\ifCLASSOPTIONcompsoc
  \section*{Acknowledgments}
\else
  \section*{Acknowledgment}
\fi

This work is supported by ARC grant DP160101458.

\ifCLASSOPTIONcaptionsoff
  \newpage
\fi



%
\balance

%

\begin{IEEEbiography}[]{Naveed Akhtar} received his Ph.D. in Computer Vision
and Pattern Recognition from the School of Computer
Science and Software Engineering, The University
of Western Australia (UWA), Australia in 2016. Previously, he did  M.Sc. in Autonomous Systems from Hochschule Bonn-Rhein-Sieg (HBRS), Germany. 
He has served as a Research Fellow with Australian National University, Australia; and UWA, and as a Research Associate with HBRS. His research
interests include Hyperspectral Imaging and Machine Learning. His work in these areas has been regularly published in prestigious research venues, including IEEE TPAMI, IEEE TNNLS, IEEE CVPR, ECCV and IEEE TGRS. He has also served as a reviewer for IEEE TPAMI, IEEE TNNLS, IEEE TIP and IEEE TGRS in the research area of hyperspectral imaging.  
\end{IEEEbiography}

\samepage
\begin{IEEEbiography}[]{Ajmal Mian}
 completed his PhD from The University
of Western Australia in 2006 with distinction
and received the Australasian Distinguished
Doctoral Dissertation Award from Computing Research
and Education Association of Australasia.
He received the prestigious Australian Postdoctoral
and Australian Research Fellowships in 2008
and 2011 respectively. He received the UWA Outstanding
Young Investigator Award in 2011, the
West Australian Early Career Scientist of the Year
award in 2012 and the Vice-Chancellors Mid-Career Research Award in
2014. He has secured seven Australian Research Council grants and one
National Health and Medical Research Council grant with a total funding
of over \$3 Million. He is currently in the School of Computer Science and
Software Engineering at The University of Western Australia and is a guest
editor of Pattern Recognition, Computer Vision and Image Understanding
and Image and Vision Computing journals. His research interests include
computer vision, machine learning, 3D shape analysis, hyperspectral image
analysis, pattern recognition, and multimodal biometrics.
\end{IEEEbiography}






\end{document}